\pdfoutput=1
\documentclass[11pt]{article}
\usepackage{authblk}
\usepackage{emnlp2021}

\usepackage{times}
\usepackage{latexsym}

\usepackage[T1]{fontenc}
\usepackage[utf8]{inputenc}

\usepackage{microtype}

\usepackage{amssymb}
\usepackage{algorithm,algorithmic}
\usepackage{booktabs}
\usepackage{multirow}
\usepackage{tabularx}
\usepackage{amsmath}
\usepackage{graphicx}

\renewcommand{\thefootnote}{\fnsymbol{footnote}}
\title{Beyond Glass-Box Features: Uncertainty Quantification Enhanced Quality Estimation for Neural Machine Translation}

\author[1]{Ke Wang}
\author[2,1]{Yangbin Shi}
\author[1,*]{Jiayi Wang}
\author[1]{Yuqi Zhang}
\author[1]{Yu Zhao}
\author[2]{Xiaolin Zheng} 
\affil[1]{Alibaba Group, Hangzhou, China} 
\affil[2]{College of Computer Science, Zhejiang University} 
\affil[ ]{
\texttt{\{moyu.wk,taiwu.syb,joanne.wjy,chenwei.zyq\}@alibaba-inc.com}}
\affil[ ]{
\texttt{kongyu@taobao.com}, \texttt{xlzheng@zju.edu.cn}
}
\newcommand\blfootnote[1]{%
  \begingroup
  \renewcommand\thefootnote{}\footnote{#1}%
  \addtocounter{footnote}{-1}%
  \endgroup
}

\begin{document}
\maketitle
\blfootnote{* indicates corresponding author.}
\renewcommand{\thefootnote}{\arabic{footnote}}
\begin{abstract}
Quality Estimation (QE) plays an essential role in applications of Machine Translation (MT). Traditionally, a QE system accepts the original source text and translation from a black-box MT system as input. Recently, a few studies indicate that as a by-product of translation, QE benefits from the model and training data's information of the MT system where the translations come from, and it is called the "glass-box QE". In this paper, we extend the definition of "glass-box QE" generally to uncertainty quantification with both "black-box" and "glass-box" approaches and design several features deduced from them to blaze a new trial in improving QE's performance. We propose a framework to fuse the feature engineering of uncertainty quantification into a pre-trained cross-lingual language model to predict the translation quality. Experiment results show that our method achieves state-of-the-art performances on the datasets of WMT 2020 QE shared task.
\end{abstract}

\section{Introduction}

The emergence of Neural Machine Translation (NMT) has brought about a revolutionary change in translation technology, resulting in translation with much higher quality. Even though NMT can produce a fairly smooth translations at present, it is still not error-free. The outputs of a machine translation (MT) system must be proofread by humans in a post-editing phase, especially in those scenes with zero tolerance for translation quality, such as in the legal domain. Therefore, it is essential to find out how good or bad the translations produced by an MT system are at run-time. 

Quality estimation (QE) aims to predict the quality of a MT system's output without any access to ground-truth translation references or human intervention. QE methods have been explored broadly \cite{blatz2004confidence, specia2009estimating, specia2018findings} on WMT's benchmark QE datasets\footnote{\url{http://www.statmt.org/wmt19/qe-task.html}}. Typical top-ranked QE systems \cite{fan2019bilingual, kim2017predictor} need a large amount of parallel corpora for pre-training and in-domain translation triplets of source texts, machine translations and corresponding quality labels/scores \cite{tercom} %for example DA scores, HTER scores or word-level tags%
for QE fine-tuning. Starting from 2019, in replacement of pre-training a model from scratch, state-of-the-art (SOTA) QE systems \cite{kepler2019unbabel,transquest} have achieved better results via taking advantage of SOTA pre-trained neural network models such as mBERT \cite{devlin-etal-2019-bert} and XLM-R \cite{xlm-r} with transfer learning to QE tasks. 

\begin{figure*}[ht]
  \includegraphics[width=\linewidth]{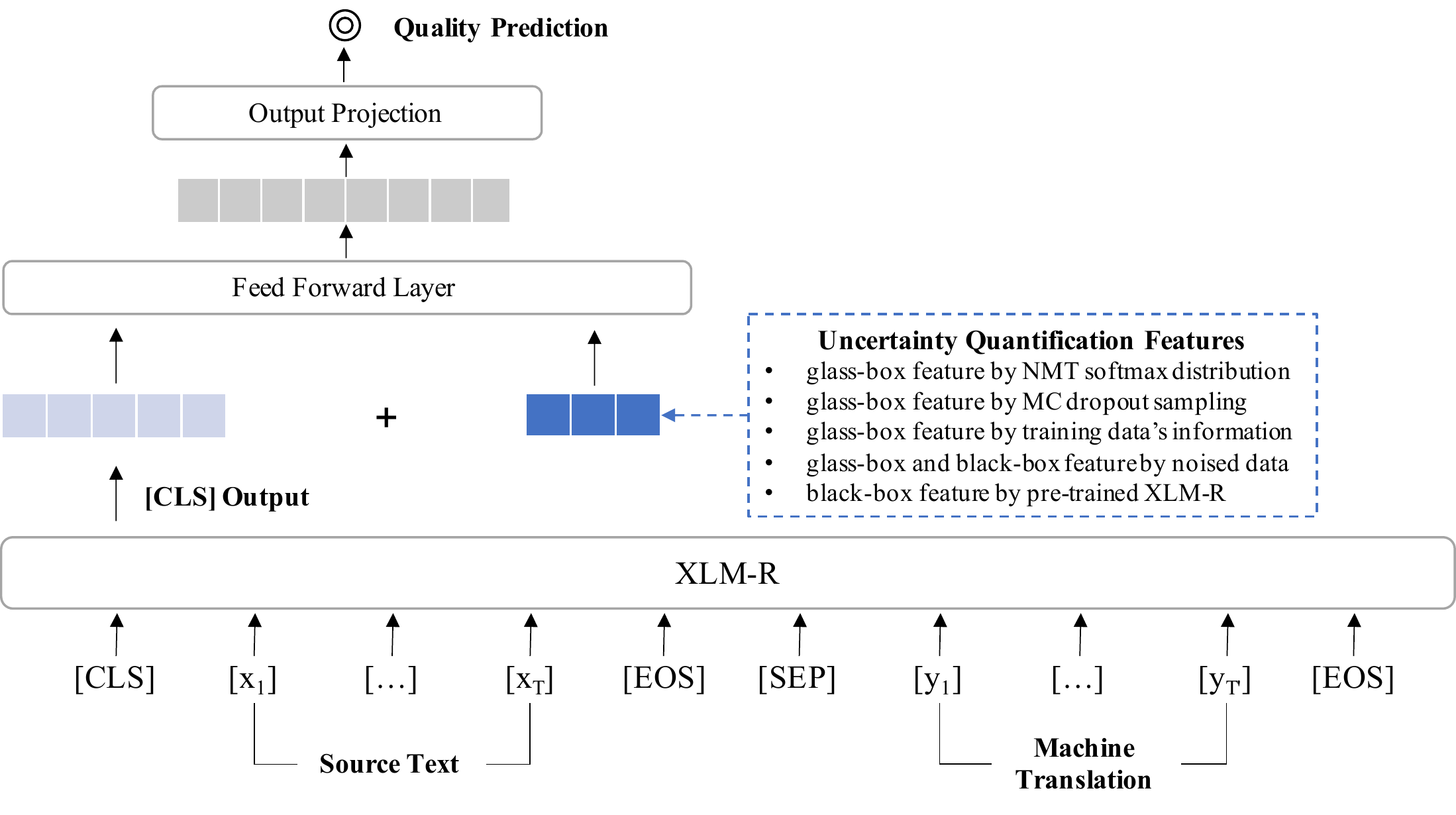}
  \caption{Structure of the uncertainty quantification feature-enhanced model.}
  \label{fig:model}
\end{figure*}

%In the previous years' WMT QE tasks before 2020, there were sub-tasks including sentence, word and document-level estimations. Since 2020, the shared task has tended to follow the human evaluation setup similar to \citet{graham-etal-2013-continuous} and released a variant of sentence-level task where the quality of machine translations is annotated with Direct Assessment (DA), instead of HTERs \cite{specia2010estimating} based on human post-editing. %At least three different raters rate the MT sentences according to a continuous scoring scale from 0 to 100 in the respect of translation quality. DA scores are standardised using the z-score by rater to be the final QE prediction.% 
%DA estimation \footnote{\url{http://www.statmt.org/wmt20/quality-estimation-task.html}} is much closer to practical applications of QE, because the human post-editing work is expensive under the actual production deployment cost control.%

In recent years, the information of the NMT systems and the corresponding training data are open to participants in WMT QE shared task, which is helpful for us to gain more QE insights. Essentially, it extends the traditional QE "black-box" NMT, where any information of the MT system is unknown to the "glass-box" stage. In fact, the concept of "glass-box" QE features has been introduced by \citet{specia2013quest}, which provides an indication of the confidence of a MT system by extracting the outputs of Moses-like Statistical Machine Translation (SMT) systems, for example the word- and phrase- alignment information and N-gram Language Model (LM) probabilities. A QE system using these features and SVM regression is considered to be the baseline model of the sentence-level tasks of WMT QE from 2013 to 2018.

Even though the LM probability, as one of the glass-box QE features, has been widely used to estimate confidence of SMT systems \cite{blatz2004confidence, specia2013quest}, the performance of using these features alone is not good enough, that can be seen from the performance of baseline results \cite{specia2013quest,QE2019findings}. For NMT's QE, the softmax output probabilities are overconfident, and it is easy to generate high confidence for points far away from the training data \cite{unsupervisedQE,kepler2019unbabel}. Therefore, it is so important to study the methods of output distribution other than 1-best prediction. 

\textit{Uncertainty Quantification}, inspired by the Bayesian framework, is representative in predicting the translation quality. A relevant method of approximation, \textit{Monte Carlo (MC) dropout} \cite{gal2016dropout}, is usually considered to be useful. Such "glass-box" methods related to MC dropout have been studied in \citet{unsupervisedQE}. Differently, we hypothesize that glass-box approaches can not only enable us to address the QE task for NMT systems in an unsupervised way, but they can enhance the black-box QE features captured from SOTA pre-trained NLP models in a supervised manner as well. On top of some efficient "glass-box" QE features, such as the expectation over predictive probabilities with MC dropout, more variants of the MC dropout sampling are exploited in our paper. As a matter of fact, our experimental results show their superiority in estimating the uncertainty of NMT models and improving the robustness of the unsupervised QE.

Translation is influenced by the source language itself \cite{zhang2019effect}. In addition to the above methods, we explicitly reduce the uncertainty quantification of the whole context of the source and machine translation to the uncertainty quantification only from the perspective of the source side. We design novel QE features obtained by both "glass-box" and "black-box" approaches to evaluate the uncertainty of source texts. Given the specific source text, these features can be easily understood as the information regarding the robustness of the NMT system and SOTA pre-trained model, jointly reflecting how difficult it is to translate the source text.

In short, our main contributions are: 
(i) we propose several novel unsupervised approaches to construct "glass-box" QE features to quantify the uncertainty of the source and machine translations, (ii) evaluate the contribution of each QE feature to the model, and finally (iii) these "glass-box" features are combined with the "black-box" QE features extracted from the pre-trained model, XLM-R, resulting in SOTA performances on the benchmark WMT 2020 QE DA datasets for 6 language pairs with different levels of training data resources.

\begin{table*}[ht]
\begin{tabular}{@{}clcccccc@{}}
\toprule
\multirow{2}{*}{Group} & \multicolumn{1}{c}{\multirow{2}{*}{Model}} & \multicolumn{2}{c}{Low Resource}  & \multicolumn{2}{c}{Mid Resource}  & \multicolumn{2}{c}{High Resource} \\
                       & \multicolumn{1}{c}{}                       & Si-En           & Ne-En           & Et-En           & Ro-En           & En-De           & En-Zh           \\ \midrule
No Feature             & TransQuest Single Model                    & 0.6365 	      & 0.7488          & 0.7437          & 0.8890          & 0.4419          & 0.4990          \\ \midrule
I                      & +$P_{step}$                                & \textbf{0.6695} & \textbf{0.7841} & \textbf{0.7654} & \textbf{0.8915} & \textbf{0.4319} & \textbf{0.5101} \\ \midrule
\multirow{3}{*}{II}    & +MC-Sim                                    & 0.6574          & \textbf{0.7926} & 0.7635          & \textbf{0.8953} & 0.4595          & 0.4917          \\
                       & +MC-Sim-Inner                              & 0.6141          & 0.7791          & 0.7607          & 0.8878          & \textbf{0.4676} & \textbf{0.5056} \\
                       & +MC-$P_{step}$                             & \textbf{0.6600} & 0.7800          & \textbf{0.7710} & 0.8905          & 0.4663          & 0.5041          \\ \midrule
\multirow{2}{*}{III}   & +DS-gram                                   & \textbf{0.6652} & \textbf{0.7859} & \textbf{0.7712} & \textbf{0.8942} & 0.4374          & 0.4980          \\
                       & +DS-neighbors                              & 0.6598          & 0.7791          & 0.7663          & 0.8840          & \textbf{0.4627} & \textbf{0.5101} \\ \midrule
\multirow{12}{*}{IV}   & +Noise-Sim-Simple                          & 0.6677          & 0.7807          & \textbf{0.7718} & 0.8887          & 0.4461          & 0.4977          \\
                       & +Noise-Sim-Simple-y                        & 0.6446          & 0.7761          & 0.7544          & 0.8864          & 0.4177          & 0.4770          \\
                       & +Noise-Sim-PE                              & 0.6478          & 0.7820          & 0.7509          & 0.8939 & 0.4370          & 0.4927          \\
                       & +Noise-Sim-PE-y                            & 0.6624          & 0.7880          & 0.7472          & 0.8921          & 0.4286          & \textbf{0.5104} \\
                       & +Noise-Sim-Inner-Simple                    & 0.6512          & 0.7778          & 0.7697          & \textbf{0.8955} & \textbf{0.4637} & 0.4522          \\
                       & +Noise-Sim-Inner-Simple-y                  & 0.6664          & \textbf{0.7912} & 0.7630          & 0.8931          & 0.4274          & 0.4960          \\
                       & +Noise-Sim-Inner-PE                        & \textbf{0.6714} & 0.7825          & 0.7463 & 0.8921          & 0.4487          & 0.4897          \\
                       & +Noise-Sim-Inner-PE-y                      & 0.6606          & 0.7787          & 0.7576          & 0.8921          & 0.4399          & 0.4909          \\
                       & +Noise-$P_{step}$-Simple                   & 0.6475          & 0.7673          & 0.7709          & 0.8916 & 0.2543          & 0.5091          \\
                       & +Noise-$P_{step}$-Simple-y                 & 0.6613          & 0.7819          & 0.7588          & 0.8899          & 0.4260          & 0.5015          \\
                       & +Noise-$P_{step}$-PE                       & 0.5615          & 0.7758          & 0.7661          & \textbf{0.8955} & 0.4300          & 0.4794          \\
                       & +Noise-$P_{step}$-PE-y                     & 0.6701          & 0.7798          & 0.7628          & 0.8953          & 0.4207          & 0.4949          \\ \midrule
\multirow{6}{*}{V}     & +MLM-$P_{mask}$-Simple                     & 0.6611          & 0.7792          & 0.7526          & 0.8885          & \textbf{0.4360} & \textbf{0.5124} \\
                       & +MLM-$P_{mask}$-Simple-y                   & 0.6410          & 0.7737          & \textbf{0.7650} & 0.8930          & 0.4187          & 0.5042          \\
                       & +MLM-$P_{mask}$-PE                         & \textbf{0.6745} & 0.7719          & 0.7552          & 0.8932          & 0.4117          & 0.4899          \\
                       & +MLM-$P_{mask}$-PE-y                       & 0.6617          & 0.7770          & 0.7629          & 0.8931          & 0.1430          & 0.5051          \\
                       & +MLM-$FP_{mask}$                           & 0.6617          & \textbf{0.7831} & 0.7639          & \textbf{0.8946} & 0.1344          & \textbf{0.4858} \\
                       & +MLM-$FP_{mask}$-y                         & 0.6560          & 0.7829          & 0.7600          & 0.8898          & 0.4141          & 0.4880          \\ \bottomrule
\end{tabular}
\caption{Pearson correlations between QE performances of our single uncertainty feature-enhanced models and human DA judgments on development sets of WMT 2020. The baseline that we compare with is the single model of the winner system in WMT 2020 QE DA task, and the results are shown in the row of "No Feature" group. Features in rows I-V are described in Section \ref{method:tp}-\ref{method:masked} respectively. Results of best models in each row are marked in bold}.
\label{tb_singlexlm}
\end{table*}
\section{Related Work}
In the previous years' WMT QE tasks before 2020, there were sub-tasks including sentence, word and document-level estimations. Since 2020, the shared task has tended to follow the human evaluation setup similar to \citet{graham-etal-2013-continuous} and released a variant of sentence-level task where the quality of machine translations is annotated with Direct Assessment (DA), instead of HTERs \cite{specia2010estimating} based on human post-editing. At least three different raters rate the MT sentences according to a continuous scoring scale from 0 to 100 in the respect of translation quality. DA scores are standardised using the z-score by rater to be the final QE prediction.
DA estimation \footnote{\url{http://www.statmt.org/wmt20/quality-estimation-task.html}} is much closer to practical applications of QE, because the human post-editing work is expensive under the actual production deployment cost control. Therefore, our experiment designs are focusing on the Direct Assessment tasks. 

Most of previous work on QE is based on the studies of feature engineering that explore how to extract useful features as inputs from source and machine translations to estimate the translation quality by a feature-enriched model. Such early work on QE, for instance, uses manually crafted features extracted from source and machine translations, and some "glass-box" features from SMT systems to build SVM regression models with RBF kernel \cite{specia2013quest}. As time went by, with the development of deep learning applied in NLP, Predictor-Estimator architecture using neural networks for QE was proposed \cite{kim2017predictor}, which relies highly on a pre-trained word prediction model with a bidirectional RNN structure, called \textit{Predictor}. Its training requires large amount of parallel data. The latent representations generated by the word prediction model are treated as features to be fed into a downstream \textit{Estimator} mode for QE fine-tuning. Motivated by Transformer \cite{transformer} framework of neural machine translation, \citet{fan2019bilingual} modified the RNN-based word predictor with a bidirectional Transformer structure to make an improvement on word prediction, leading to better QE results.

SOTA pre-trained models have achieved successes in various NLP tasks \cite{mbert,xlm-r} with transfer leaning. Current SOTA QE systems \cite{ist-unbabel,kepler2019unbabel,transquest}, profit from SOTA pre-trained models to gain cross-lingual representations of source and machine translations and fine tune the model with additional layers to meet the goals of QE. They passingly remove the dependency of large amount of parallel data and ease the burden of pre-training complex neural network models. Encouraged by the top ranked QE system \cite{transquest}, we design a feature-enhanced model similar to their work, that also relies on the pre-trained XLM-R model, but it is enhanced by a mixture of other useful "glass-box" and "black-box" QE features. The model structure can be first glimpsed in Figure \ref{fig:model} and details will be introduced in Section \ref{method}.

% \usepackage{booktabs}
% \usepackage{multirow}
\begin{table*}[ht]
\scriptsize
\begin{tabular}{@{}cccccc@{}}
\toprule
Sort by                                                                                            & Languages & 1st                    & 2nd                      & 3rd                       & 4th                       \\ \midrule
\multirow{6}{*}{\begin{tabular}[c]{@{}c@{}}Performance of\\      the enhanced model\end{tabular}}      & Si-En     & MLM-$P_{mask}$-PE      & Noise-Sim-Inner-PE       & Noise-$P_{step}$-PE-y     & $P_{step}$                \\
                                                                                                   & Ne-En     & MC-Sim                 & Noise-Sim-Inner-Simple-y & Noise-Sim-PE-y            & DS-gram                   \\
                                                                                                   & Et-En     & Noise-Sim-Simple       & DS-gram                  & MC-$P_{step}$             & Noise-$P_{step}$-Simple   \\
                                                                                                   & Ro-En     & Noise-Sim-Inner-Simple & Noise-$P_{step}$-PE      & MC-Sim                    & Noise-$P_{step}$-PE-y     \\
                                                                                                   & En-De     & MC-Sim-Inner           & MC-$P_{step}$            & Noise-Sim-Inner-Simple    & DS-neighbors              \\
                                                                                                   & En-Zh     & MLM-$P_{mask}$-Simple  & Noise-Sim-PE-y           & DS-neighbors              & $P_{step}$                \\ \midrule
\multirow{6}{*}{\begin{tabular}[c]{@{}c@{}}Correlations to\\      human DA judgments\end{tabular}} & Si-En     & MC-$P_{step}$          & MC-Sim-Inner             & $P_{step}$                & MC-Sim                    \\
                                                                                                   & Ne-En     & MC-$P_{step}$          & $P_{step}$               & Noise-$P_{step}$-Simple   & Noise-$P_{step}$-PE       \\
                                                                                                   & Et-En     & MC-$P_{step}$          & MC-Sim-Inner             & MC-Sim                    & $P_{step}$                \\
                                                                                                   & Ro-En     & MC-$P_{step}$          & $P_{step}$               & Noise-$P_{step}$-Simple-y & Noise-$P_{step}$-Simple   \\
                                                                                                   & En-De     & MC-$P_{step}$          & $P_{step}$               & Noise-$P_{step}$-Simple   & Noise-$P_{step}$-Simple-y \\
                                                                                                   & En-Zh     & $P_{step}$             & MC-$P_{step}$            & Noise-$P_{step}$-Simple   & MC-Sim-Inner              \\ \bottomrule
\end{tabular}
\caption{Most useful uncertainty features for each language pair. }
\label{tab:top_feature}
\end{table*}

\section{Methodology}
\label{method}
In this section, we provide a complete view of our uncertainty quantification approaches: 
(1) the predictive information of softmax distribution from the NMT model is still used as a "glass-box" QE feature due to its indication in QE explored from previous work \cite{ist-unbabel}. We describe it simply in Section \ref{method:tp};
(2) Stimulated by \citet{unsupervisedQE}, more useful derivatives of MC dropout sampling for uncertainty quantification are investigated as "glass-box" features in Section \ref{method:mc}; 
(3) we extend the meaning of "glass-box" in a broader sense and shift our gaze from model confidence to data confidence in Section \ref{method:ngram}. More creatively, (4) a combination of "glass-box" and "black-box" approaches is proposed to estimate the uncertainty of source texts in Section \ref{method:noise}. In particular, (5) the "black-box" approach utilizing a SOTA pre-trained NLP model in (4) can inherently estimate the confidence of the sources via the masking strategy in Section \ref{method:masked}. In Section \ref{method:model}, a model enhanced by above uncertainty features is carried out for the final goal of quality estimation. 
\subsection{Quantify uncertainty with softmax distribution of NMT model}
\label{method:tp}
For auto-regressive sequence generating models like Transformers \cite{transformer}, decoding probability at each step can be extracted from the softmax layer directly in a "glass-box" setting:
\begin{equation}
    P_{step}^{(\mathbf{x},t,\theta)}=\log P(y_t|\mathbf{y}_{<t},\mathbf{x},\theta)
\end{equation}
where $\mathbf{x}$ represents the input source text and $\mathbf{y}$ is the output machine translation.
$P_{step}$ is a probability sequence with the same length of the generated sequence $\mathbf{y}$. Three statistical indicators of the sequence can be used to estimate uncertainty of the output: the expectation, standard deviation, and the combined ratio of them:
\begin{equation}
    \mathbb{E}(P_{step}|\mathbf{x},\theta)=\frac{1}{T}\sum_{t=1}^{T}{P_{step}^{(\mathbf{x},t,\theta)}}
\end{equation}
\begin{equation}
\begin{aligned}
    & \sigma(P_{step}|\mathbf{x},\theta)\\
    =& \sqrt{\mathbb{E}(P_{step}^2|\mathbf{x},\theta)-\mathbb{E}^2(P_{step}|\mathbf{x},\theta)}
\end{aligned}
\end{equation}
\begin{equation}
    Combo(P_{step}|\mathbf{x},\theta)=\frac{\mathbb{E}(P_{step}|\mathbf{x},\theta)}{\sigma(P_{step}|\mathbf{x},\theta)}
\end{equation}
In general, higher probability expectation and lower probability variance usually indicate that the model is more confident about the output.
$P_{step}$ is an extended version of the $TP$ feature in \citet{unsupervisedQE} and the expectation of $P_{step}$ is the same as $TP$. In our feature-enhanced model, when we mention the feature $P_{step}$, it actually means a vector of the three statistical indicators rather than a single value $TP$. The same way is applied to other features in the following sub-sections.

\subsection{Quantify uncertainty with Monte Carlo Dropout}
\label{method:mc}
Monte Carlo Dropout \cite{gal2016dropout} is an efficient "glass-box" approach to estimate uncertainty. It enables random dropout on neural networks during inference to obtain measures of uncertainty. Output sequences $\hat{\mathbf{y}}$ sampled across stochastic forward-passes by MC dropout with sampled model parameters $\hat{\theta}$ can be different. Intuitively, if $\mathbf{y}$ is a high-quality output with small uncertainty, the Monte Carlo sampled outputs $\hat{\mathbf{y}}$ should be similar to $\mathbf{y}$ and the diversity among them should be low. Hence, two measurements of sampling based on text similarity are carried out here:
\begin{equation}
    MC\textit{-}Sim=Sim(\mathbf{y}, \hat{\mathbf{y}})
\end{equation}
\begin{equation}
    MC\textit{-}Sim\textit{-}Inner=Sim(\hat{\mathbf{y}}_i,\hat{\mathbf{y}}_j)
\end{equation}
For the similarity score function, as in \citet{unsupervisedQE}, Meteor metric \cite{meteor} is applied.

Besides, as a sentence-level probability score, $\mathbb{E}(P_{step})$ can also be calculated with different model parameters $\hat{\theta}$ by MC dropout sampling:
\begin{equation}
    MC\textit{-}P_{step}=\mathbb{E}(P_{step}|\mathbf{x},\hat{\theta})
\end{equation}
The expectation, standard deviation, and combined ratio of $MC\textit{-}Sim$, $MC\textit{-}Sim\textit{-}Inner$ and $MC\textit{-}P_{step}$ are calculated over all MC dropout samples and will be used as "glass-box" uncertainty quantification features. Among them, $\mathbb{E}(MC\textit{-}P_{step})$, $\sigma(MC\textit{-}P_{step})$, $Combo(MC\textit{-}P_{step})$, and $\mathbb{E}(MC\textit{-}Sim\textit{-}Inner)$ are equivalent to $D\textit{-}TP$, $D\textit{-}Var$, $D\textit{-}Combo$, and $D\textit{-}Lex\textit{-}Sim$ in \citet{unsupervisedQE}

\subsection{Quantify uncertainty with informative training data}
\label{method:ngram}
For "glass-box" QE, not only the NMT model is helpful, but the information of the training data is valuable as well. A simple but widely-used feature, the rate of N-grams of the source text covered by the NMT training data, is used and defined as follows:
\begin{equation}
\begin{aligned}
 & DS\textit{-}gram^{(N)} \\
 = & \frac{\sharp(\{\mathbf{x}_{i\leq t<i+N} |\mathbf{x}_{i\leq t<i+N} \in \textit{ train sets}\})}{T-N+1} \cdot 
\end{aligned}
\end{equation}
We consider $N$ from 1 to 5 for N-grams in the coverage rate calculation. This "glass-box" feature measures how the source text to be translated is far away from the model's training data, thereby quantifying how confident the NMT model is to produce the corresponding machine translation. 

The above N-gram feature is widely used in SMT's QE, but is not strong enough for NMT. Inspired by the idea of k-nearest-neighbor machine translation \cite{knnmt}, if the similarity between the input $\mathbf{x}$ and nearest neighbors from the train sets is relatively high, the NMT model tends to produce a high-quality output. Instead of complex calculation in \citet{knnmt}, we propose a simple data-level "glass-box" feature based on data similarity for uncertainty quantification:
\begin{equation}
\begin{aligned}
    & DS\textit{-}neighbors\textit{-}x^{(K)} \\
    =& \frac{1}{K}\sum_{k=1}^K{Sim(\mathbf{x},\mathbf{x'}^{(k)})}
\end{aligned}
\end{equation}
% \begin{equation}
% \begin{aligned}
%     & DS\textit{-}neighbors\textit{-}y^{(K)} \\
%     =& \frac{1}{K}\sum_{k=1}^K{Sim(\mathbf{y},\mathbf{y'}^{(k)})}
% \end{aligned}
% \end{equation}
where $\mathbf{x'}^{(k)}$ is the $k$-th nearest neighbor of $\mathbf{x}$ in train sets according to the Levenshtein Distance. Simultaneously, $DS\textit{-}neighbors\textit{-}y$ is defined similarly. $DS\textit{-}neighbors\textit{-}x$ measures how familiar the NMT model is with the input $\mathbf{x}$, while $DS\textit{-}neighbors\textit{-}y$ measures how fluent the output $\mathbf{y}$ is based on observation of training data. 
\subsection{Quantify uncertainty with noised data}
\label{method:noise}
Monte Carlo Dropout approaches in \ref{method:mc} can be regarded as a robustness test with noise in the model. Due to its validity in \citet{unsupervisedQE}, it is rational to believe that a similar way with appropriate noise in the input of MT will perform comparably. 

Therefore, we define the following "glass-box" uncertainty quantification measures similar to those in \ref{method:mc}. The only difference is that the NMT model weights are fixed $\theta$ without MC dropout sampling and the model decodes translations $\tilde{\mathbf{y}}$ with a noised input $\tilde{\mathbf{x}}$.

\begin{equation}
\label{noise_sim}
    Noise\textit{-}Sim=Sim(\mathbf{y}, \tilde{\mathbf{y}})
\end{equation}
\begin{equation}
\label{noise_dsim}
    Noise\textit{-}Sim\textit{-}Inner=Sim(\tilde{\mathbf{y}}_i,\tilde{\mathbf{y}}_j)
\end{equation}
\begin{equation}
\label{noise_tp}
    Noise\textit{-}P_{step}=\mathbb{E}(P_{step}|\tilde{\mathbf{x}},\theta)
\end{equation}

\begin{algorithm}[t]
\caption{Generate Noise Input with "Post-Editing"}
\begin{algorithmic}[1]
\label{alg:noise}
% \small
\REQUIRE input $\mathbf{x}=\{x_t|t=1,2,...,T\}$, hyper-parameters $R$, $p_i$, $p_d$.
\STATE Initialize $\mathbf{x}_{mask}=\mathbf{x}$
\FOR{$r=1,...,R$} 
	\STATE $\mathbf{x}_{mask}$ = randomly delete tokens from $\mathbf{x}_{mask}$ with probability $p_d$
	\STATE $\mathbf{x}_{mask}$ = randomly insert special \textit{<mask>} tokens into $\mathbf{x}_{mask}$ with probability $p_i$
\ENDFOR
\STATE $\tilde{\mathbf{x}}=MLM(\mathbf{x}_{mask})$, where $MLM$ is a pre-trained masked language model.
\RETURN $\tilde{\mathbf{x}}$
\end{algorithmic} 
\end{algorithm}
One crucial point in this approach is how to generate noised input $\tilde{\mathbf{x}}$. One solution is a "black-box" way that utilizes the masking strategy of pre-trained multi-lingual NLP models. Basically, we can mask some words in the source text and get a noised source text by the prediction of the pre-trained model in the masked positions. In implementation, we mask each source token $x_t$ successively and obtain the predictions $\tilde{x_t}$ from a pre-trained masked language model to gain a set of noised source texts ${\tilde{\mathbf{x}}}$. This simple approach only performs substitution on $\mathbf{x}$, but limits the diversity of the noised samples.

\citet{catape} proposed an automatic post-editing algorithm which imitates post-editing process of human post-editing via constructing atomic operations including insertion, deletion, and substitution. \citet{humanlabel} also applied a similar algorithm for QE's data augmentation. In our case, adding appropriate noise to input data is a "post-editing" process on input $\mathbf{x}$.
To enrich the noise space of $\mathbf{x}$, we adjust the imitation learning algorithm in \citet{catape} to a simplified version to obtain noised input ${\tilde{\mathbf{x}}}$. We "post-edit" the input $\mathbf{x}$ by randomly deleting tokens and inserting masks for several rounds to get $\mathbf{x_m}$. Then, a SOTA pre-trained model predicts the tokens in the masked positions of $\mathbf{x_m}$
to get the post-edited $\tilde{\mathbf{x}}$. Pseudo codes of this "post-editing" algorithm is provided in Algorithm \ref{alg:noise}. 

The two methods of noised data acquisition mentioned above both involve the task of masked token prediction. In a sense, the translation $\mathbf{y}$ can be appended with $\mathbf{x_m}$ and fed into the pre-trained multilingual model as a semantic constraint to obtain a more reasonable prediction. Hence there are four variants for features in Equation \ref{noise_sim} to \ref{noise_tp}: "simple" approach and "post-edit" approach, each one can be with or without $\mathbf{y}$ during masked tokens prediction. In the rest of the paper, they are denoted like $Noise\textit{-}Sim\textit{-}Simple$, $Noise\textit{-}Sim\textit{-}PE$, $Noise\textit{-}Sim\textit{-}Simple\textit{-}y$, and $Noise\textit{-}Sim\textit{-}PE\textit{-}y$ respectively.

\subsection{Quantify uncertainty with pre-trained model}
\label{method:masked}
In Algorithm \ref{alg:noise}, the pre-trained masked language model is used to predict masked tokens in $\mathbf{x_m}$. In this process, similar to $P_{step}$ in NMT model, the prediction probability of each masked token can be extracted.
\begin{equation}
\begin{aligned}
MLM\textit{-}P_{mask}\textit{-}Simple=\log{P({\tilde{x_t}}|\mathbf{x}_{\neq t})}
\end{aligned}
\end{equation}
\begin{equation}
\begin{aligned}
\text{\small{$MLM\textit{-}P_{mask}\textit{-}Simple\textit{-}y$}}=\log{P({\tilde{x_t}}|\mathbf{x}_{\neq t},\mathbf{y})}
\end{aligned}
\end{equation}
\begin{equation}
MLM\textit{-}P_{mask}\textit{-}PE=\log{P(\tilde{x_t}|\mathbf{x}_m)}
\end{equation}
\begin{equation}
MLM\textit{-}P_{mask}\textit{-}PE\textit{-}y=\log{P(\tilde{x_t}|\mathbf{x}_m,\mathbf{y})}
\end{equation}

For the simple approach in Algorithm \ref{alg:noise}, not only the top-1 probability can be extracted as $MLM\textit{-}P_{mask}$, but the forced decoding probability can also be extracted from the softmax distribution:
\begin{equation}
MLM\textit{-}FP_{mask}=\log{P({x_t}|\mathbf{x}_{\neq t})}
\end{equation}
\begin{equation}
MLM\textit{-}FP_{mask}\textit{-}y=\log{P({x_t}|\mathbf{x}_{\neq t},\mathbf{y})}
\end{equation}
Different from the "glass-box" feature $P_{step}$, $MLM\textit{-}P_{mask}$ and $MLM\textit{-}FP_{mask}$ are "black-box" features since they does not require any access to the NMT model. Instead, knowledge from the pre-trained model is the key to measure uncertainty in these features. 

\subsection{Uncertainty feature-enhanced model}
\label{method:model}
All uncertainty features proposed in Section \ref{method:tp}-\ref{method:masked} can be regarded as unsupervised approaches for quality estimation. Even if unsupervised approaches do not require human QE labeling, their performances are still far below those of supervised approaches with transfer learning on SOTA pre-trained models \cite{bergamot,transquest}.

Some previous work has explored combining "glass-box" QE features with transfer learning ways and achieved top results \cite{bergamot,ist-unbabel}. We design a feature-enhanced model framework with transfer learning as well, reusing a multi-lingual pre-trained NLP model, XLM-R \cite{xlm-r} that assists to achieve the top-ranked QE results in WMT 2020 QE DA task. We concatenate the source text and machine translation and feed them into the pre-trained XLM-R to get the output representation of the special [CLS] token. Afterwards, it is concatenated with multiple normalized uncertainty features proposed in Section \ref{method:tp}-\ref{method:masked}, and fed into a simple linear regression layer to predict the translation quality score. The effectiveness of proposed uncertainty quantification features can also be evaluated according to the model's performance. The architecture of the uncertainty quantification feature-enhanced model is shown in Figure \ref{fig:model}.

\section{Experiments}

\subsection{Setup}
\label{sec:setup}
\textbf{Dataset.} The MLQE dataset proposed by \citet{mlqe} is used for the WMT2020 QE shared tasks. We evaluate our work on this open public dataset and compare our model with SOTA QE system on the DA task. To explore the performance of our model on different languages, we conduct all experiments on 6 language pairs with different levels of NMT training data resources: English-German (En–De) and  English-Chinese (En-Zh) for high-resource ones, Romanian-English (Ro-En) and Estonian-English (Et-En) for midum-resource ones, and Nepali-English (Ne-En) and Sinhala-English (Si-En) for low-resource ones.
% We choice DA rather than HTER as the golden-standard quality labels because DA judgements are available for all six language pairs on the dataset while HTER scores are only available for En-De and En-Zh. 
% \usepackage{booktabs}
% \usepackage{multirow}
\begin{table}[ht]
\small
\setlength{\tabcolsep}{5pt}
\begin{tabular}{@{}ccccccc@{}}
\toprule
\multirow{2}{*}{$k$} & \multicolumn{2}{c}{Low Resource}  & \multicolumn{2}{c}{Mid Resource}  & \multicolumn{2}{c}{High Resource} \\
                     & Si-En           & Ne-En           & Et-En           & Ro-En           & En-De           & En-Zh           \\ \midrule
0                    & 0.6365 	       & 0.7488          & 0.7437          & 0.8890          & 0.4419          & 0.4990         \\
1                    & 0.6745          & 0.7926          & 0.7718          & 0.8955          & 0.4676          & 0.5124          \\
2                    & 0.6597          & 0.7782          & 0.7587          & 0.8985          & 0.4501          & 0.4934          \\
3                    & 0.6602          & 0.7888          & 0.7618          & \textbf{0.9019} & 0.4727          & 0.5095          \\
4                    & \textbf{0.6808} & 0.7693          & 0.7680          & 0.9003          & 0.4619          & 0.5055          \\
5                    & 0.6622          & 0.7788          & 0.7623          & 0.8907          & 0.4274          & \textbf{0.5522} \\
6                    & 0.6677          & 0.7740          & 0.7674          & 0.8918          & 0.4523          & 0.5210          \\
7                    & 0.6621          & 0.7785          & 0.7603          & 0.8981          & 0.4570          & 0.5135          \\
8                    & 0.6461          & 0.7839          & \textbf{0.7818} & 0.8947          & 0.4252          & 0.4620          \\
9                    & 0.6714          & 0.7889          & 0.7725          & 0.8995          & 0.4403          & 0.5325          \\
10                   & 0.6682          & 0.7802          & 0.7558          & 0.8870          & 0.4693          & 0.5401          \\
11                   & 0.6614          & 0.7814          & 0.7748          & 0.8992          & 0.4531          & 0.5029          \\
12                   & 0.6703          & \textbf{0.7937} & 0.7651          & 0.8956          & 0.4198          & 0.5119          \\
13                   & 0.6663          & 0.7876          & 0.6512          & 0.8979          & 0.4251          & 0.5274          \\
14                   & 0.6701          & 0.7747          & 0.7693          & 0.8967          & 0.4341          & 0.4890          \\
15                   & 0.6663          & 0.7804          & 0.7700          & 0.9000          & 0.4690          & 0.5298          \\
16                   & 0.6516          & 0.7804          & 0.7665          & 0.8981          & 0.4271          & 0.5116          \\
17                   & 0.6659          & 0.7750          & 0.7562          & 0.9011          & 0.4260          & 0.5185          \\
18                   & 0.6624          & 0.7876          & 0.7514          & 0.9008          & 0.4105          & 0.5179          \\
19                   & 0.6676          & 0.7780          & 0.7737          & 0.8989          & 0.4450          & 0.5154          \\
20                   & 0.6559          & 0.7632          & 0.7630          & 0.8973          & 0.4203          & 0.5016          \\
21                   & 0.6456          & 0.7877          & 0.7664          & 0.8978          & \textbf{0.4823} & 0.5105          \\
22                   & 0.6645          & 0.7612          & 0.7514          & 0.8909          & 0.1726          & 0.5326          \\
23                   & 0.6750          & 0.7758          & 0.7469          & 0.8889          & 0.2487          & 0.5287          \\
24                   & 0.6677          & 0.7733          & 0.7626          & 0.8905          & 0.2684          & 0.5140          \\ \bottomrule
\end{tabular}
\caption{Pearson correlations between QE results of top-$k$ uncertainty feature-enhanced models and the ground-truth DA labels on the development sets.}
\label{tab:topk}
\end{table}

\textbf{Baseline.} In this paper, we mainly focus on improvement on single model from uncertainty quantification features. Models with strategies including ensemble and data augmentation are not listed for comparing as these strategies can also be applied to our model. The transfer learning model based on XLM-R from \citet{transquest} is selected as a strong baseline, as TransQuest is the winner of the WTM20 QE DA task in all language pairs and the code of this model is released with detailed hyper-parameters\footnote{\url{https://github.com/TharinduDR/TransQuest}}. We set parameter n\_fold to 1 in the released code for fair comparison.

% 切入点：glass-box特征不能单看pearson，要考虑它给pre-train model带来的增益，而部分高pearson的特征可能与hidden state的overlap比较大，因此给模型带来的增益不一定大

\begin{table*}[ht]
\begin{tabular}{@{}ccccccc@{}}
\toprule
\multicolumn{1}{c}{\multirow{2}{*}{Model}} & \multicolumn{2}{c}{Low-Resource}                      & \multicolumn{2}{c}{Mid-Resource}                      & \multicolumn{2}{c}{High-Resource}                     \\
\multicolumn{1}{c}{}                       & \multicolumn{1}{c}{Si-En} & \multicolumn{1}{c}{Ne-En} & \multicolumn{1}{c}{Et-En} & \multicolumn{1}{c}{Ro-En} & \multicolumn{1}{c}{En-De} & \multicolumn{1}{c}{En-zh} \\ \midrule
OpenKiwi (Official Baseline)               & 0.3737                    & 0.3860                    & 0.4770                    & 0.6845                    & 0.1455                    & 0.1902                    \\
Transquest's Single Model                  & 0.6207                    & 0.7641                    & 0.7386                    & 0.8812                    & 0.3772                    & 0.4715                    \\
Our Singe Feature Enhanced   Model         & 0.6607                    & 0.7954                    & 0.7950                    & 0.8948                    & 0.4774                    & 0.4969                    \\
Our Multiple Features Enhanced   Model     & \textbf{0.6677}           & \textbf{0.7980}           & \textbf{0.8021}           & \textbf{0.8986}           & \textbf{0.5086}           & \textbf{0.5242}           \\ \bottomrule
\end{tabular}
\caption{Final QE results on the test sets of WMT 2020 QE DA task.}
\label{tab:final}
\end{table*}
\subsection{Useful features for different languages}
All the uncertainty quantification features can be evaluated directly by calculating the Pearson correlation with the ground-truth labels as in \citet{unsupervisedQE}. In this way, each feature extractor can be regarded as a unsupervised model.  We put the results in Appendix \ref{sec:appendix} because unsupervised method is not the key point in this paper. Besides, these features are not completely "orthogonal" to the representation from the pre-trained model. In another word, part of the information from uncertainty quantification features is already covered by the pre-trained multi-lingual language model. Feature with higher Pearson correlation does not necessarily mean higher performance when combined with the "black-box" QE features from the pre-trained model.
Therefore, in our uncertainty quantification enhanced model, the importance of the QE features should be evaluated by the performance increment after incorporating them into the model.

We concatenate each group of normalized uncertainty features with the outputs of the [CLS] token, and then fine-tune the pre-trained XLM-R model with a simple linear regression layer as shown in Figure \ref{fig:model}. The Pearson correlations of each enhanced model on the development sets are listed in Table \ref{tb_singlexlm}. We summarize the most useful features for each language pair according to the performance of enhanced model with the single feature and the feature's correlation with human DA scores without model fine tuning separately in Table \ref{tab:top_feature}.

From the results, we can conclude that 1) In most cases, uncertainty quantification feature-enhanced model outperforms the non-feature baseline. 2) The performance gains from a feature are various for different language pairs. For example, $DS\textit{-}neighbors$ enhanced model achieves higher performance in high-resource languages, while $Noise-Sim$ features work better on low-resource languages. 3) Sometimes, features with high correlation to human DA scores may not be good ones for the enhanced model, as the information comes from these features might have been covered by the pre-trained model already.

\subsection{Multiple feature-enhanced results}
Experiments on each single feature above provide us guidance to enhance model with multiple features. We sort all the uncertainty quantification features for each language pairs according to the performance of single feature-enhanced model in Table \ref{tb_singlexlm}. Then we conduct experiments with top $k$ features of each language pair. Results with different values of $k$ on the development sets in Table \ref{tab:topk} indicate that our uncertainty feature-enhanced model can be further improved with multiple groups of features.

Finally, based on the results on the development sets, we select the most appropriate features for each language pair and predict QE scores on the test sets with the multiple feature-enhanced model.
% We also conduct Williams Significance Test \cite{williamsTest} to compare with the baseline. 
The final results on the test sets in Table \ref{tab:final} show that our uncertainty feature-enhanced model outperforms the official baseline \cite{specia2020findings} and the model of TransQuest, which is the winner of the WMT2020 QE shared task of DA.

\section{Conclusion}
In this paper, we extend previous work on "glass-box" QE for uncertainty quantification and explore how SOTA transfer learning method can benefit from uncertainty features. First, we re-organize "glass-box" features from the softmax distribution and Monte Carlo Dropout sampling in previous work and derive useful variants. Secondly, based on the information of training data of the NMT model, the "glass-box" features in the respect of data attributes are extracted to predict uncertainty as well. More importantly, we propose a new method, which utilizes the masking mechanism of the pre-trained model to quantify uncertainty through robustness testing via several pre-designed "glass-box" features. Finally, we evaluate all the "black-box" and "glass-box" approaches by an uncertainty feature enhanced model on the benchmark DA datasets of WMT 2020 QE shared task. The experimental results show that our proposed features for  uncertainty estimation are effective, and the uncertainty feature-enhanced QE model is superior to SOTA QE systems.

\section*{Acknowledgements}
This work is supported by National Key R\&D Program of China (2018YFB1403202).

\bibliography{anthology,custom}

\begin{thebibliography}{27}
\expandafter\ifx\csname natexlab\endcsname\relax\def\natexlab#1{#1}\fi

\bibitem[{Blatz et~al.(2004)Blatz, Fitzgerald, Foster, Gandrabur, Goutte,
  Kulesza, Sanchis, and Ueffing}]{blatz2004confidence}
John Blatz, Erin Fitzgerald, George Foster, Simona Gandrabur, Cyril Goutte,
  Alex Kulesza, Alberto Sanchis, and Nicola Ueffing. 2004.
\newblock Confidence estimation for machine translation.
\newblock In \emph{Coling 2004: Proceedings of the 20th international
  conference on computational linguistics}, pages 315--321.

\bibitem[{Conneau et~al.(2019)Conneau, Khandelwal, Goyal, Chaudhary, Wenzek,
  Guzm{\'a}n, Grave, Ott, Zettlemoyer, and Stoyanov}]{xlm-r}
Alexis Conneau, Kartikay Khandelwal, Naman Goyal, Vishrav Chaudhary, Guillaume
  Wenzek, Francisco Guzm{\'a}n, Edouard Grave, Myle Ott, Luke Zettlemoyer, and
  Veselin Stoyanov. 2019.
\newblock Unsupervised cross-lingual representation learning at scale.
\newblock \emph{arXiv preprint arXiv:1911.02116}.

\bibitem[{Denkowski and Lavie(2014)}]{meteor}
Michael Denkowski and Alon Lavie. 2014.
\newblock Meteor universal: Language specific translation evaluation for any
  target language.
\newblock In \emph{Proceedings of the ninth workshop on statistical machine
  translation}, pages 376--380.

\bibitem[{Devlin et~al.(2019)Devlin, Chang, Lee, and
  Toutanova}]{devlin-etal-2019-bert}
Jacob Devlin, Ming-Wei Chang, Kenton Lee, and Kristina Toutanova. 2019.
\newblock \href {https://doi.org/10.18653/v1/N19-1423} {{BERT}: Pre-training of
  deep bidirectional transformers for language understanding}.
\newblock In \emph{Proceedings of the 2019 Conference of the North {A}merican
  Chapter of the Association for Computational Linguistics: Human Language
  Technologies, Volume 1 (Long and Short Papers)}, pages 4171--4186,
  Minneapolis, Minnesota. Association for Computational Linguistics.

\bibitem[{Fan et~al.(2019)Fan, Wang, Li, Zhou, Chen, and Si}]{fan2019bilingual}
Kai Fan, Jiayi Wang, Bo~Li, Fengming Zhou, Boxing Chen, and Luo Si. 2019.
\newblock “bilingual expert” can find translation errors.
\newblock In \emph{Proceedings of the AAAI Conference on Artificial
  Intelligence}, volume~33, pages 6367--6374.

\bibitem[{Fomicheva et~al.(2020{\natexlab{a}})Fomicheva, Sun, Fonseca, Blain,
  Chaudhary, Guzm{\'a}n, Lopatina, Specia, and Martins}]{mlqe}
Marina Fomicheva, Shuo Sun, Erick Fonseca, Fr{\'e}d{\'e}ric Blain, Vishrav
  Chaudhary, Francisco Guzm{\'a}n, Nina Lopatina, Lucia Specia, and
  Andr{\'e}~FT Martins. 2020{\natexlab{a}}.
\newblock Mlqe-pe: A multilingual quality estimation and post-editing dataset.
\newblock \emph{arXiv preprint arXiv:2010.04480}.

\bibitem[{Fomicheva et~al.(2020{\natexlab{b}})Fomicheva, Sun, Yankovskaya,
  Blain, Chaudhary, Fishel, Guzm{\'a}n, and Specia}]{bergamot}
Marina Fomicheva, Shuo Sun, Lisa Yankovskaya, Fr{\'e}d{\'e}ric Blain, Vishrav
  Chaudhary, Mark Fishel, Francisco Guzm{\'a}n, and Lucia Specia.
  2020{\natexlab{b}}.
\newblock Bergamot-latte submissions for the wmt20 quality estimation shared
  task.
\newblock In \emph{Proceedings of the Fifth Conference on Machine Translation},
  pages 1010--1017.

\bibitem[{Fomicheva et~al.(2020{\natexlab{c}})Fomicheva, Sun, Yankovskaya,
  Blain, Guzm{\'a}n, Fishel, Aletras, Chaudhary, and Specia}]{unsupervisedQE}
Marina Fomicheva, Shuo Sun, Lisa Yankovskaya, Fr{\'e}d{\'e}ric Blain, Francisco
  Guzm{\'a}n, Mark Fishel, Nikolaos Aletras, Vishrav Chaudhary, and Lucia
  Specia. 2020{\natexlab{c}}.
\newblock Unsupervised quality estimation for neural machine translation.
\newblock \emph{Transactions of the Association for Computational Linguistics},
  8:539--555.

\bibitem[{Fonseca et~al.(2019)Fonseca, Yankovskaya, Martins, Fishel, and
  Federmann}]{QE2019findings}
Erick Fonseca, Lisa Yankovskaya, Andr{\'e}~FT Martins, Mark Fishel, and
  Christian Federmann. 2019.
\newblock Findings of the wmt 2019 shared tasks on quality estimation.
\newblock In \emph{Proceedings of the Fourth Conference on Machine Translation
  (Volume 3: Shared Task Papers, Day 2)}, pages 1--10.

\bibitem[{Gal and Ghahramani(2016)}]{gal2016dropout}
Yarin Gal and Zoubin Ghahramani. 2016.
\newblock Dropout as a bayesian approximation: Representing model uncertainty
  in deep learning.
\newblock In \emph{international conference on machine learning}, pages
  1050--1059. PMLR.

\bibitem[{Graham et~al.(2013)Graham, Baldwin, Moffat, and
  Zobel}]{graham-etal-2013-continuous}
Yvette Graham, Timothy Baldwin, Alistair Moffat, and Justin Zobel. 2013.
\newblock \href {https://www.aclweb.org/anthology/W13-2305} {Continuous
  measurement scales in human evaluation of machine translation}.
\newblock In \emph{Proceedings of the 7th Linguistic Annotation Workshop and
  Interoperability with Discourse}, pages 33--41, Sofia, Bulgaria. Association
  for Computational Linguistics.

\bibitem[{Kepler et~al.(2019)Kepler, Tr{\'e}nous, Treviso, Vera, G{\'o}is,
  Farajian, Lopes, and Martins}]{kepler2019unbabel}
Fabio Kepler, Jonay Tr{\'e}nous, Marcos Treviso, Miguel Vera, Ant{\'o}nio
  G{\'o}is, M~Amin Farajian, Ant{\'o}nio~V Lopes, and Andr{\'e}~FT Martins.
  2019.
\newblock Unbabel's participation in the wmt19 translation quality estimation
  shared task.
\newblock \emph{arXiv preprint arXiv:1907.10352}.

\bibitem[{Khandelwal et~al.(2020)Khandelwal, Fan, Jurafsky, Zettlemoyer, and
  Lewis}]{knnmt}
Urvashi Khandelwal, Angela Fan, Dan Jurafsky, Luke Zettlemoyer, and Mike Lewis.
  2020.
\newblock Nearest neighbor machine translation.
\newblock \emph{arXiv preprint arXiv:2010.00710}.

\bibitem[{Kim et~al.(2017)Kim, Jung, Kwon, Lee, and Na}]{kim2017predictor}
Hyun Kim, Hun-Young Jung, Hongseok Kwon, Jong-Hyeok Lee, and Seung-Hoon Na.
  2017.
\newblock Predictor-estimator: Neural quality estimation based on target word
  prediction for machine translation.
\newblock \emph{ACM Transactions on Asian and Low-Resource Language Information
  Processing (TALLIP)}, 17(1):1--22.

\bibitem[{Moura et~al.(2020)Moura, Vera, van Stigt, Kepler, and
  Martins}]{ist-unbabel}
Joao Moura, Miguel Vera, Daan van Stigt, Fabio Kepler, and Andr{\'e}~FT
  Martins. 2020.
\newblock Ist-unbabel participation in the wmt20 quality estimation shared
  task.
\newblock In \emph{Proceedings of the Fifth Conference on Machine Translation},
  pages 1029--1036.

\bibitem[{Pires et~al.(2019)Pires, Schlinger, and Garrette}]{mbert}
Telmo Pires, Eva Schlinger, and Dan Garrette. 2019.
\newblock How multilingual is multilingual bert?
\newblock In \emph{Proceedings of the 57th Annual Meeting of the Association
  for Computational Linguistics}, pages 4996--5001.

\bibitem[{Ranasinghe et~al.(2020)Ranasinghe, Orǎsan, and Mitkov}]{transquest}
Tharindu Ranasinghe, Constantin Orǎsan, and Ruslan Mitkov. 2020.
\newblock Transquest at wmt2020: Sentence-level direct assessment.
\newblock In \emph{Proceedings of the Fifth Conference on Machine Translation},
  pages 1049--1055.

\bibitem[{Snover et~al.(2005)Snover, Dorr, Schwartz, Makhoul, Micciulla, and
  Weischedel}]{tercom}
Mathew Snover, Bonnie Dorr, Richard Schwartz, John Makhoul, Linnea Micciulla,
  and Ralph Weischedel. 2005.
\newblock A study of translation error rate with targeted human annotation.
\newblock In \emph{Proceedings of the 7th Conference of the Association for
  Machine Translation in the Americas (AMTA 06)}, pages 223--231.

\bibitem[{Specia et~al.(2020)Specia, Blain, Fomicheva, Fonseca, Chaudhary,
  Guzm{\'a}n, and Martins}]{specia2020findings}
Lucia Specia, Fr{\'e}d{\'e}ric Blain, Marina Fomicheva, Erick Fonseca, Vishrav
  Chaudhary, Francisco Guzm{\'a}n, and Andr{\'e}~FT Martins. 2020.
\newblock Findings of the wmt 2020 shared task on quality estimation.
\newblock In \emph{Proceedings of the Fifth Conference on Machine Translation},
  pages 743--764.

\bibitem[{Specia et~al.(2018)Specia, Blain, Logacheva, Astudillo, and
  Martins}]{specia2018findings}
Lucia Specia, Fr{\'e}d{\'e}ric Blain, Varvara Logacheva, Ram{\'o}n Astudillo,
  and Andr{\'e} Martins. 2018.
\newblock Findings of the wmt 2018 shared task on quality estimation.
\newblock Association for Computational Linguistics.

\bibitem[{Specia and Farzindar(2010)}]{specia2010estimating}
Lucia Specia and Atefeh Farzindar. 2010.
\newblock Estimating machine translation post-editing effort with hter.
\newblock In \emph{Proceedings of the Second Joint EM+/CNGL Workshop Bringing
  MT to the User: Research on Integrating MT in the Translation Industry (JEC
  10)}, pages 33--41.

\bibitem[{Specia et~al.(2013)Specia, Shah, De~Souza, and
  Cohn}]{specia2013quest}
Lucia Specia, Kashif Shah, Jos{\'e}~GC De~Souza, and Trevor Cohn. 2013.
\newblock Quest-a translation quality estimation framework.
\newblock In \emph{Proceedings of the 51st Annual Meeting of the Association
  for Computational Linguistics: System Demonstrations}, pages 79--84.

\bibitem[{Specia et~al.(2009)Specia, Turchi, Cancedda, Dymetman, and
  Cristianini}]{specia2009estimating}
Lucia Specia, Marco Turchi, Nicola Cancedda, Marc Dymetman, and Nello
  Cristianini. 2009.
\newblock Estimating the sentence-level quality of machine translation systems.
\newblock In \emph{13th Conference of the European Association for Machine
  Translation}, pages 28--37.

\bibitem[{Tuan et~al.(2021)Tuan, El-Kishky, Renduchintala, Chaudhary,
  Guzm{\'a}n, and Specia}]{humanlabel}
Yi-Lin Tuan, Ahmed El-Kishky, Adithya Renduchintala, Vishrav Chaudhary,
  Francisco Guzm{\'a}n, and Lucia Specia. 2021.
\newblock Quality estimation without human-labeled data.
\newblock In \emph{Proceedings of the 16th Conference of the European Chapter
  of the Association for Computational Linguistics: Main Volume}, pages
  619--625.

\bibitem[{Vaswani et~al.(2017)Vaswani, Shazeer, Parmar, Uszkoreit, Jones,
  Gomez, Kaiser, and Polosukhin}]{transformer}
Ashish Vaswani, Noam Shazeer, Niki Parmar, Jakob Uszkoreit, Llion Jones,
  Aidan~N Gomez, Lukasz Kaiser, and Illia Polosukhin. 2017.
\newblock Attention is all you need.
\newblock In \emph{NIPS}.

\bibitem[{Wang et~al.(2020)Wang, Wang, Ge, Shi, Zhao, and Fan}]{catape}
Ke~Wang, Jiayi Wang, Niyu Ge, Yangbin Shi, Yu~Zhao, and Kai Fan. 2020.
\newblock Computer assisted translation with neural quality estimation and
  auotmatic post-editing.
\newblock In \emph{Proceedings of the 2020 Conference on Empirical Methods in
  Natural Language Processing: Findings}, pages 2175--2186.

\bibitem[{Zhang and Toral(2019)}]{zhang2019effect}
Mike Zhang and Antonio Toral. 2019.
\newblock The effect of translationese in machine translation test sets.
\newblock \emph{arXiv preprint arXiv:1906.08069}.

\end{thebibliography}
\bibliographystyle{acl_natbib}

% \clearpage
\appendix

\section{Uncertainty features as unsupervised approach to QE}
\label{sec:appendix}
All uncertainty quantification features proposed in Section \ref{method:tp}-\ref{method:masked} can be regarded as a unsupervised approach to quality estimation. The Pearson correlations between each component of these features and the human annotated ground-truth labels are calculated in Table \ref{tb_singlepearson_1} and Table \ref{tb_singlepearson_2}.

% \usepackage{booktabs}
% \usepackage{multirow}
\begin{table*}[htp]
\setlength{\tabcolsep}{3pt}
\begin{tabular}{@{}cllllllll@{}}
\toprule
\multirow{2}{*}{Group} & \multicolumn{1}{c}{\multirow{2}{*}{Feature}} & \multicolumn{1}{c}{\multirow{2}{*}{Component}} & \multicolumn{2}{c}{Low Resource}  & \multicolumn{2}{c}{Mid Resource}  & \multicolumn{2}{c}{High Resource} \\
                       & \multicolumn{1}{c}{}                         & \multicolumn{1}{c}{}                           & si-en           & ne-en           & et-en           & ro-en           & en-de           & en-zh           \\ \midrule
\multirow{36}{*}{IV}   & Noise-Sim-Simple                             & $E$                                            & 0.2362          & 0.2327          & 0.3468          & 0.3443          & 0.1286          & 0.1697          \\
                       & Noise-Sim-Simple                             & $Std$                                          & 0.1834          & 0.2197          & 0.3242          & 0.4093          & 0.0947          & 0.1966          \\
                       & Noise-Sim-Simple                             & $Combo$                                        & 0.0532          & 0.0247          & 0.2321          & 0.1652          & 0.0703          & 0.0079          \\
                       & Noise-Sim-Simple-y                           & $E$                                            & 0.2293          & 0.2834          & 0.3060          & 0.3990          & 0.1257          & 0.1463          \\
                       & Noise-Sim-Simple-y                           & $Std$                                          & 0.1984          & 0.3174          & 0.3557          & 0.5027          & 0.0894          & 0.1698          \\
                       & Noise-Sim-Simple-y                           & $Combo$                                        & 0.0110          & 0.0487          & 0.1685          & 0.2474          & 0.0595          & 0.0152          \\
                       & Noise-Sim-PE                                 & $E$                                            & 0.2554          & 0.3118          & 0.3272          & 0.4801          & 0.0961          & 0.1980          \\
                       & Noise-Sim-PE                                 & $Std$                                          & 0.1885          & 0.0694          & 0.2120          & 0.0228          & 0.0759          & 0.0059          \\
                       & Noise-Sim-PE                                 & $Combo$                                        & 0.0173          & 0.1551          & 0.3927          & 0.3692          & 0.1388          & 0.1545          \\
                       & Noise-Sim-PE-y                               & $E$                                            & 0.3123          & 0.3503          & 0.3818          & 0.5014          & 0.0877          & 0.2067          \\
                       & Noise-Sim-PE-y                               & $Std$                                          & 0.1807          & 0.1179          & 0.2621          & 0.0326          & 0.0802          & 0.0385          \\
                       & Noise-Sim-PE-y                               & $Combo$                                        & 0.0477          & 0.1518          & 0.4439          & 0.4107          & 0.1579          & 0.1447          \\
                       & Noise-Sim-Inner-Simple                       & $E$                                            & 0.3557          & 0.3213          & 0.3903          & 0.4119          & 0.0953          & 0.2368          \\
                       & Noise-Sim-Inner-Simple                       & $Std$                                          & 0.2078          & 0.2190          & 0.3062          & 0.4311          & 0.0947          & 0.2141          \\
                       & Noise-Sim-Inner-Simple                       & $Combo$                                        & 0.1731          & 0.1861          & 0.3615          & 0.1357          & 0.0655          & 0.1917          \\
                       & Noise-Sim-Inner-Simple-y                     & $E$                                            & \textbf{0.3871} & 0.4179          & 0.3681          & 0.4990          & 0.0779          & 0.0669          \\
                       & Noise-Sim-Inner-Simple-y                     & $Std$                                          & 0.2708          & 0.3160          & 0.3674          & 0.5303          & 0.0957          & 0.0647          \\
                       & Noise-Sim-Inner-Simple-y                     & $Combo$                                        & 0.3069          & 0.3600          & 0.3601          & 0.4129          & 0.0758          & 0.1518          \\
                       & Noise-Sim-Inner-PE                           & $E$                                            & 0.2859          & 0.2654          & 0.3106          & 0.4612          & 0.0256          & 0.1895          \\
                       & Noise-Sim-Inner-PE                           & $Std$                                          & 0.2145          & 0.1078          & 0.0572          & 0.1836          & 0.0489          & 0.0066          \\
                       & Noise-Sim-Inner-PE                           & $Combo$                                        & 0.0955          & 0.0429          & 0.1707          & 0.0461          & 0.1074          & 0.0590          \\
                       & Noise-Sim-Inner-PE-y                         & $E$                                            & 0.3321          & 0.3000          & 0.3787          & 0.4791          & 0.0252          & 0.0967          \\
                       & Noise-Sim-Inner-PE-y                         & $Std$                                          & 0.1947          & 0.1218          & 0.0569          & 0.1777          & 0.0424          & 0.0246          \\
                       & Noise-Sim-Inner-PE-y                         & $Combo$                                        & 0.0836          & 0.0341          & 0.2378          & 0.1133          & 0.1205          & 0.0370          \\
                       & Noise-$P_{step}$-Simple                      & $E$                                            & 0.3639          & \textbf{0.4866} & \textbf{0.4870} & 0.6449          & \textbf{0.2046} & \textbf{0.2630} \\
                       & Noise-$P_{step}$-Simple                      & $Std$                                          & 0.3089          & 0.2755          & 0.2955          & 0.3751          & 0.1217          & 0.2182          \\
                       & Noise-$P_{step}$-Simple                      & $Combo$                                        & 0.1546          & 0.1125          & 0.0348          & 0.0115          & 0.0039          & 0.0535          \\
                       & Noise-$P_{step}$-Simple-y                    & $E$                                            & 0.3661          & 0.4798          & 0.4841          & \textbf{0.6465} & 0.1870          & 0.2576          \\
                       & Noise-$P_{step}$-Simple-y                    & $Std$                                          & 0.3298          & 0.3226          & 0.3289          & 0.4495          & 0.1648          & 0.2125          \\
                       & Noise-$P_{step}$-Simple-y                    & $Combo$                                        & 0.1310          & 0.0530          & 0.0374          & 0.0058          & 0.0339          & 0.0106          \\
                       & Noise-$P_{step}$-PE                          & $E$                                            & 0.2978          & 0.4643          & 0.4046          & 0.5820          & 0.0845          & 0.1910          \\
                       & Noise-$P_{step}$-PE                          & $Std$                                          & 0.2233          & 0.2450          & 0.0530          & 0.2973          & 0.0703          & 0.0493          \\
                       & Noise-$P_{step}$-PE                          & $Combo$                                        & 0.0182          & 0.0309          & 0.2562          & 0.1735          & 0.0025          & 0.1149          \\
                       & Noise-$P_{step}$-PE-y                        & $E$                                            & 0.3179          & 0.4544          & 0.4010          & 0.5718          & 0.0654          & 0.1970          \\
                       & Noise-$P_{step}$-PE-y                        & $Std$                                          & 0.2678          & 0.2582          & 0.0322          & 0.2432          & 0.0865          & 0.0944          \\
                       & Noise-$P_{step}$-PE-y                        & $Combo$                                        & 0.0526          & 0.0098          & 0.2159          & 0.1961          & 0.0060          & 0.0744          \\ \midrule
\end{tabular}
\caption{(PART-I) Pearson correlations between all single uncertainty quantification features and human DA judgments on development sets of WMT 2020 QE DA task. Features in group IV are described in Section \ref{method:noise}. Each feature has multiple components including expectation ($E$), standard deviation ($Std$) and a combined ratio of the two ($Combo$). Results of best models in each group are marked in bold. Considering some features have negative correlation with the labels, the absolute values are remained.}.
\label{tb_singlepearson_1}
\end{table*}

\begin{table*}[]
\setlength{\tabcolsep}{3pt}
\begin{tabular}{@{}cllllllll@{}}
\toprule
\multirow{2}{*}{Group} & \multicolumn{1}{c}{\multirow{2}{*}{Feature}} & \multicolumn{1}{c}{\multirow{2}{*}{Component}} & \multicolumn{2}{c}{Low Resource}  & \multicolumn{2}{c}{Mid Resource}  & \multicolumn{2}{c}{High Resource} \\
                       & \multicolumn{1}{c}{}                         & \multicolumn{1}{c}{}                           & Si-En           & Ne-En           & Et-En           & Ro-En           & En-De           & En-Zh           \\ \midrule
\multirow{3}{*}{I}     & $P_{step}$                                   & $E$                                            & \textbf{0.4437} & \textbf{0.5315} & \textbf{0.4870} & \textbf{0.6481} & 0.2061          & 0.2583          \\
                       & $P_{step}$                                   & $Std$                                          & 0.4185          & 0.4724          & 0.4712          & 0.5958          & \textbf{0.2603} & \textbf{0.3020} \\
                       & $P_{step}$                                   & $Combo$                                        & 0.0430          & 0.0680          & 0.0493          & 0.0182          & 0.0941          & 0.0234          \\ \midrule
\multirow{9}{*}{II}    & MC-$P_{step}$                                & $E$                                            & \textbf{0.4819} & \textbf{0.5429} & \textbf{0.6199} & \textbf{0.6956} & 0.2157          & \textbf{0.2979} \\
                       & MC-$P_{step}$                                & $Std$                                          & 0.4285          & 0.3848          & 0.3473          & 0.4191          & \textbf{0.2636} & 0.2592          \\
                       & MC-$P_{step}$                                & $Combo$                                        & 0.1962          & 0.1767          & 0.1620          & 0.0120          & 0.1089          & 0.0344          \\
                       & MC-Sim                                       & $E$                                            & 0.4007          & 0.4230          & 0.4981          & 0.6257          & 0.1407          & 0.2473          \\
                       & MC-Sim                                       & $Std$                                          & 0.1323          & 0.1560          & 0.0023          & 0.1080          & 0.0324          & 0.0719          \\
                       & MC-Sim                                       & $Combo$                                        & 0.2419          & 0.3063          & 0.4823          & 0.4686          & 0.1332          & 0.2078          \\
                       & MC-Sim-Inner                                 & $E$                                            & 0.4451          & 0.4617          & 0.5165          & 0.6248          & 0.1760          & 0.2610          \\
                       & MC-Sim-Inner                                 & $Std$                                          & 0.2098          & 0.1724          & 0.0117          & 0.2857          & 0.0648          & 0.1124          \\
                       & MC-Sim-Inner                                 & $Combo$                                        & 0.2388          & 0.3221          & 0.5447          & 0.4806          & 0.1661          & 0.2034          \\ \midrule
\multirow{15}{*}{III}  & DS-gram                                  & 1-gram                                         & \textbf{0.1451} & \textbf{0.1783} & \textbf{0.4201} & \textbf{0.4557} & 0.0812          & 0.1277          \\
                       & DS-gram                                  & 2-gram                                         & 0.1446          & 0.1585          & 0.3605          & 0.4227          & 0.0453          & \textbf{0.1871} \\
                       & DS-gram                                  & 3-gram                                         & 0.0541          & 0.0502          & 0.2746          & 0.3227          & 0.0579          & 0.1141          \\
                       & DS-gram                                  & 4-gram                                         & 0.0336          & 0.0119          & 0.1954          & 0.2362          & 0.0071          & 0.0727          \\
                       & DS-gram                                  & 5-gram                                         & 0.0507          & 0.0089          & 0.1019          & 0.1565          & 0.0147          & 0.0596          \\
                       & DS-neighbors                                 & 1 neighbor of $\mathbf{x}$                     & 0.0860          & 0.0416          & 0.1723          & 0.3208          & 0.1289          & 0.1139          \\
                       & DS-neighbors                                 & 3 neighbor of $\mathbf{x}$                     & 0.0469          & 0.0550          & 0.1681          & 0.2566          & 0.0899          & 0.1283          \\
                       & DS-neighbors                                 & 5 neighbors of $\mathbf{x}$                    & 0.0324          & 0.0654          & 0.1608          & 0.2044          & 0.0756          & 0.1232          \\
                       & DS-neighbors                                 & 10 neighbors of $\mathbf{x}$                   & 0.0347          & 0.0699          & 0.1431          & 0.1711          & 0.0677          & 0.0972          \\
                       & DS-neighbors                                 & 30 neighbors of $\mathbf{x}$                   & 0.0325          & 0.0794          & 0.1049          & 0.0736          & 0.0983          & 0.0208          \\
                       & DS-neighbors                                 & 1 neighbor of $\mathbf{y}$                     & 0.0331          & 0.0349          & 0.1006          & 0.1332          & 0.1699          & 0.1297          \\
                       & DS-neighbors                                 & 3 neighbor of $\mathbf{y}$                     & 0.0210          & 0.0489          & 0.1222          & 0.1495          & 0.1763          & 0.1496          \\
                       & DS-neighbors                                 & 5 neighbors of $\mathbf{y}$                    & 0.0150          & 0.0464          & 0.1098          & 0.1572          & 0.1769          & 0.1569          \\
                       & DS-neighbors                                 & 10 neighbors of $\mathbf{y}$                   & 0.0115          & 0.0399          & 0.1060          & 0.1597          & \textbf{0.1773} & 0.1625          \\
                       & DS-neighbors                                 & 30 neighbors of $\mathbf{y}$                   & 0.0207          & 0.0447          & 0.1089          & 0.1548          & 0.1733          & 0.1409          \\ \midrule
\multirow{18}{*}{V}    & MLM-$P_{mask}$-Simple                        & $E$                                            & 0.1905          & 0.0996          & 0.3332          & 0.1254          & 0.0431          & 0.1490          \\
                       & MLM-$P_{mask}$-Simple                        & $Std$                                          & 0.1154          & 0.0546          & 0.2797          & 0.0891          & 0.0008          & 0.0811          \\
                       & MLM-$P_{mask}$-Simple                        & $Combo$                                        & 0.1993          & 0.0995          & 0.2659          & 0.1337          & 0.0719          & 0.1666          \\
                       & MLM-$P_{mask}$-Simple-y                      & $E$                                            & \textbf{0.3511} & \textbf{0.3707} & 0.3619          & 0.4402          & 0.0498          & 0.1698          \\
                       & MLM-$P_{mask}$-Simple-y                      & $Std$                                          & 0.2933          & 0.3102          & 0.3210          & 0.2818          & 0.0044          & 0.0856          \\
                       & MLM-$P_{mask}$-Simple-y                      & $Combo$                                        & 0.3039          & 0.3187          & \textbf{0.3866} & \textbf{0.4473} & 0.0757          & 0.1888          \\
                       & MLM-$P_{mask}$-PE                            & $E$                                            & 0.1874          & 0.1506          & 0.3214          & 0.1414          & 0.0429          & 0.1680          \\
                       & MLM-$P_{mask}$-PE                            & $Std$                                          & 0.1289          & 0.0739          & 0.0121          & 0.0915          & 0.1006          & 0.0403          \\
                       & MLM-$P_{mask}$-PE                            & $Combo$                                        & 0.1978          & 0.1388          & 0.2590          & 0.1852          & \textbf{0.1435} & 0.0704          \\
                       & MLM-$P_{mask}$-PE-y                          & $E$                                            & 0.2966          & 0.3641          & 0.3816          & 0.4142          & 0.0779          & 0.2033          \\
                       & MLM-$P_{mask}$-PE-y                          & $Std$                                          & 0.0982          & 0.0144          & 0.0598          & 0.0115          & 0.0613          & 0.0146          \\
                       & MLM-$P_{mask}$-PE-y                          & $Combo$                                        & 0.2512          & 0.2005          & 0.3087          & 0.2671          & 0.1216          & 0.1341          \\
                       & MLM-$FP_{mask}$                              & $E$                                            & 0.2241          & 0.1708          & 0.3263          & 0.1552          & 0.0780          & \textbf{0.2203} \\
                       & MLM-$FP_{mask}$                              & $Std$                                          & 0.1587          & 0.1769          & 0.2844          & 0.1546          & 0.0659          & 0.2069          \\
                       & MLM-$FP_{mask}$                              & $Combo$                                        & 0.2012          & 0.0870          & 0.2692          & 0.0837          & 0.0578          & 0.1347          \\
                       & MLM-$FP_{mask}$-y                            & $E$                                            & 0.2978          & 0.3517          & 0.3211          & 0.3259          & 0.0575          & 0.2181          \\
                       & MLM-$FP_{mask}$-y                            & $Std$                                          & 0.1953          & 0.2745          & 0.2405          & 0.2263          & 0.0543          & 0.1994          \\
                       & MLM-$FP_{mask}$-y                            & $Combo$                                        & 0.3406          & 0.3322          & 0.3714          & 0.4298          & 0.0588          & 0.1678          \\ \midrule
\end{tabular}
\caption{(PART-II) Pearson correlations between all single uncertainty quantification features and human DA judgments on development sets of WMT 2020 QE DA task. Features in groups I,II,III and V are described in Section \ref{method:tp}, \ref{method:mc}, \ref{method:ngram}, and \ref{method:masked}.}.
\label{tb_singlepearson_2}
\end{table*}

\end{document}